\documentclass[10pt,twocolumn,letterpaper]{article}

\usepackage{cvpr}              

\usepackage{colortbl}
\usepackage{multirow}
\usepackage{booktabs}
\usepackage{amssymb}
\usepackage{amsmath}
\usepackage{bm}

\definecolor{lightcyan}{rgb}{0.88, 1.0, 1.0}
\definecolor{deemph}{gray}{0.85}
\definecolor{baselinecolor}{gray}{.9}

\definecolor{cvprblue}{rgb}{0.21,0.49,0.74}
\usepackage[pagebackref,breaklinks,colorlinks,allcolors=cvprblue]{hyperref}

\def\paperID{37959} 
\def\confName{CVPR}
\def\confYear{2026}

\title{Exploring Adaptive Masked Reconstruction for Self-Supervised \\ Skeleton-Based Action Recognition}

\author{Shengkai Sun\textsuperscript{1}, Zhiyong Cheng\textsuperscript{1}\thanks{Corresponding author: jason.zy.cheng@gmail.com} ,  Zefan Zhang\textsuperscript{2}, Jianfeng Dong\textsuperscript{3}, Zhihui Li\textsuperscript{4}, Meng Wang\textsuperscript{1}\\
\textsuperscript{1}Hefei University of Technology \ \  \textsuperscript{2}Jilin University\\
\textsuperscript{3}Zhejiang Gongshang University \ \ \textsuperscript{4}University of Science and Technology of China\\
\href{https://github.com/AshenOne1005/AMR}{https://github.com/AshenOne1005/AMR} \\
}

\begin{document}
\maketitle
\begin{abstract}
Recently, masked skeleton reconstruction models have emerged as strong action representation learners, driving significant progress in self-supervised skeleton‑based action recognition. However, existing state‑of‑the‑art methods must predict an exceedingly large number of spatiotemporal patches, significantly prolonging training time. Besides, by treating all spatiotemporal regions equally during reconstruction, these models are distracted from learning the critical motion patterns that underlie action semantics. To address these challenges, we propose Adaptive Masked Reconstruction (AMR), a faster and stronger pre‑training framework. We first decouple the decoder from the encoder, enabling flexible prediction of larger spatiotemporal patches and dramatically reducing reconstruction complexity. Given that larger patches contain more complex information, which is challenging to predict and consequently degrades performance, we accordingly introduce an adaptive guidance module. This module identifies regions of high motion informativeness, guiding the model to focus on the most discriminative parts of each patch and alleviating reconstruction difficulty. Experiments on NTU RGB+D 60, NTU RGB+D 120, and PKU-MMD datasets demonstrate that AMR not only accelerates pre‑training substantially but also improves downstream recognition accuracy, surpassing current state‑of‑the‑art approaches.
\end{abstract}    
\section{Introduction}
\label{sec:intro}
Skeleton-based action recognition \cite{yan2018spatial,cheng2020skeleton} methods represent the human body through a set of keypoints, providing a lightweight representation. This formulation not only offers an efficient data structure but also exhibits strong robustness against variations in illumination and background, thereby attracting significant attention within the human action recognition community in recent years.
Although substantial progress has been made in skeleton-based action recognition, existing methods predominantly rely on the supervised learning paradigm and consequently demand large volumes of labor-intensive annotations, which severely limit their scalability in real-world applications. In recent years, the remarkable success of self-supervised learning (SSL) in natural language processing and computer vision \cite{he2020momentum,bardes2022vicreg,xu2024deep} has spurred its adoption in the skeleton-based action recognition domain \cite{zheng2018unsupervised,dong2023hierarchical,sun2023unified}. This paradigm shift significantly reduces the reliance on annotated data. Typically, a two-stage training procedure is employed: during the pre-training stage, the model exploits the intrinsic structure of the skeleton sequences to generate supervisory signals and learn generalizable representations without manual labels; in the downstream fine-tuning stage, the pre-trained encoder is adapted for the target task by appending a classifier and performing end-to-end supervised fine-tuning.

\begin{figure}[tb!]
\centering
\scalebox{0.95}{
\includegraphics[width=0.95\columnwidth]{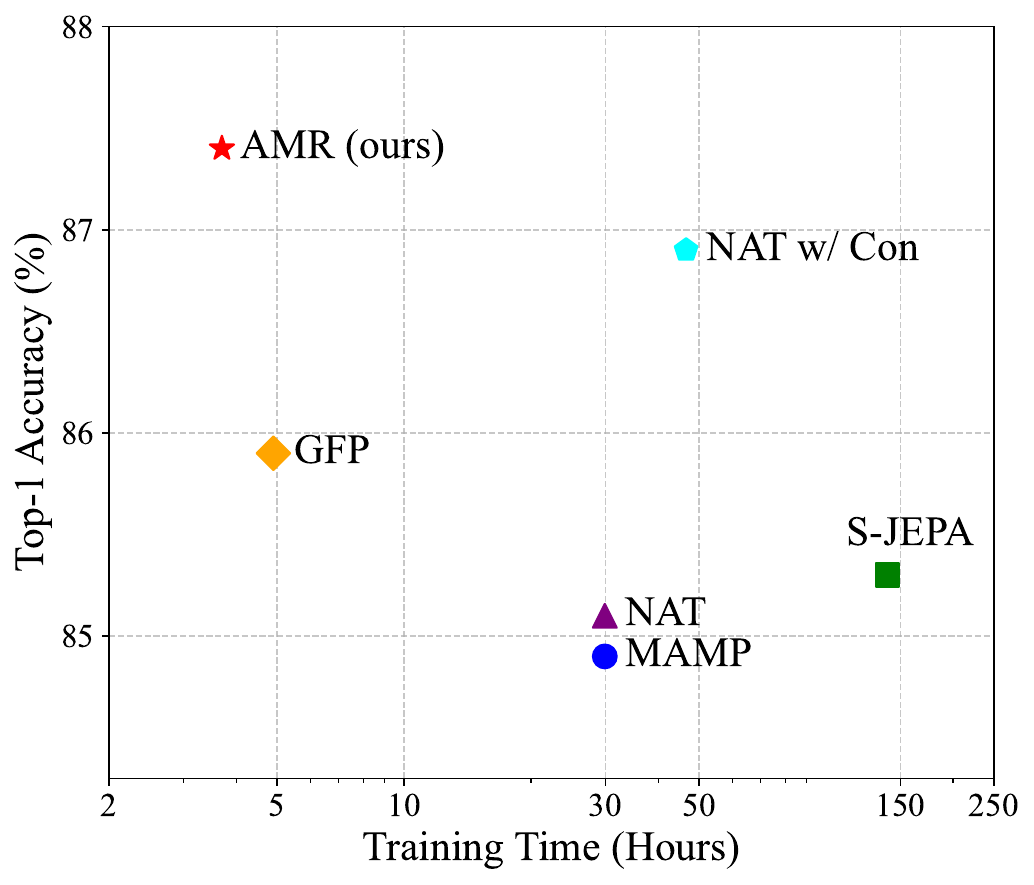}
}
\vspace{-3mm}
\caption{Top-1 accuracy of linear evaluation and training time of different masked skeleton reconstruction methods (NTU-60 x-sub). 
}\label{fig:intro_fig}
\end{figure}

Early self-supervised methods \cite{yang2023self} for skeleton-based action recognition primarily focused on designing handcrafted pretext tasks (\eg, motion prediction \cite{kim2022global}, joint ordering \cite{lin2020ms2l}). However, the learned representations exhibited limited generalization. Current research efforts predominantly center on two paradigms: contrastive learning \cite{guo2022contrastive,zhang2022contrastive} and masked skeleton reconstruction \cite{mao2023masked}. In the contrastive learning paradigm, diverse augmentations (\eg, spatial transformations, temporal cropping) are applied to the input sequence to form positive pairs, while other samples serve as negatives. An instance discrimination objective then encourages the model to learn highly discriminative features. Masked skeleton reconstruction methods draw inspiration from the masked autoencoder (MAE) \cite{he2022masked} used in visual representation learning. Here, a skeleton motion sequence is divided into multiple spatiotemporal patches, each capturing the movement of specific joints over a defined spatial and temporal region. A large proportion of these patches are randomly masked, compelling the model to reconstruct the missing data solely from the visible contextual cues. This process effectively drives the model to capture both the holistic structural patterns and the underlying semantic information of the skeleton sequence, yielding rich and robust representations. Note that the MAE framework is naturally compatible with the Vision Transformer (ViT) \cite{dosovitskiy2020image} architecture, fully leveraging the Transformer’s strength in modeling long-range spatiotemporal dependencies. Consequently, recent state‑of‑the‑art methods based on masked skeleton reconstruction typically demonstrate superior performance.

Despite the impressive performance of existing skeleton MAEs \cite{mao2023masked,sun2025towards}, several non-trivial challenges persist. First, although Transformers excel at contextual modeling, their computational complexity remains high, particularly for long sequences.
Skeletal action data inherently comprise many consecutive frames, often forming ultra-long input sequences when partitioned into numerous patches, which dramatically increases the computational burden. 
An intuitive solution is to reconstruct larger patches to shorten the sequence. However, this strategy substantially degrades model performance. The underlying reason is that larger patches encapsulate far more complex and diverse patterns, making it difficult for the model to effectively capture their intricate internal spatiotemporal dependencies. Nevertheless, this performance degradation \textit{can} be mitigated. This stems from a second critical limitation in the current paradigm: it applies uniform reconstruction targets to all masked regions regardless of their semantic importance. Crucially, understanding action semantics relies on capturing regions that contain critical movement cues. Consequently, treating all regions equally in reconstruction forces the model to expend its limited modeling capacity on redundant information, which not only increases reconstruction difficulty but also undermines effective modeling of the essential motion dynamics, ultimately leading to degraded performance.

To address the issues above, we propose a novel self-supervised pre-training approach, \textbf{A}daptive \textbf{M}asked \textbf{R}econstruction (\textbf{AMR}). The core innovation of AMR lies in its integration of \textbf{decoder decoupling} and \textbf{focal reconstruction}. Specifically, we decouple the decoder from the encoder. This architectural decoupling enables flexible prediction of larger patches while significantly reducing computational complexity. 
Conventional decoding concatenates features of visible patches with learnable mask tokens before feeding them into a Transformer decoder. This process involves computationally expensive self-attention among mask tokens and cross-attention between mask tokens and visible patch features. We observe that the self-attention computation among mask tokens is redundant. Relying solely on cross-attention with visible patch features yields sufficiently context-aware representations.
Accordingly, we replace the standard self-attention decoder with an efficient cross-attention decoder. This module employs a set of learnable \textit{query} vectors as decoder input, while the visible patch features serve simultaneously as \textit{keys} and \textit{values}. By simply adjusting the number of queries (corresponding to the desired count and granularity of target patches), the architecture allows flexible prediction of patches of varying sizes. 
Furthermore, we introduce a simple yet effective mechanism, termed focal reconstruction, to mitigate the issue of performance degradation in large patch reconstruction. 
Its core principle is to adaptively guide the model to reduce focus on unimportant, redundant regions and concentrate modeling efforts on motion-critical areas essential for action semantics.
We first quantify motion energy within each local spatiotemporal window. Based on the quantified motion energy, we design a well-behaved scaling function. This function generates a weighting map to implement adaptive focal reconstruction, effectively reducing reconstruction difficulty and encouraging the model to learn more discriminative feature representations.
Consequently, our model maintains performance comparable to small patch reconstruction while significantly outperforming conventional masked reconstruction methods, establishing a self-supervised pre-training paradigm that is faster to train and delivers superior recognition performance. As shown in Figure \ref{fig:intro_fig}, we achieve a favorable trade-off between computational efficiency and model performance.

In summary, our main contributions are:
\begin{itemize}
    \item We decouple the encoder and decoder in the skeleton MAE and design a decoder module relying on cross-attention. This design significantly reduces computational complexity and supports flexible large patch prediction, substantially decreasing the time cost of masked reconstruction pre-training.
    \item We introduce a focal reconstruction strategy guided by the quantification of local motion energy, which adaptively directs the model's focus to regions containing critical motion information. This mechanism effectively mitigates the performance degradation associated with increasing patch size.
    \item We conduct extensive experiments on three widely-used skeleton-based action recognition benchmarks. The results demonstrate the effectiveness of our proposed method, showcasing significant superiority over current state-of-the-art methods in both training efficiency and downstream task recognition performance.
\end{itemize}

\section{Related Work}
\label{sec:related_work}

\subsection{Self-Supervised Learning for Skeleton-Based Action}
Self-supervised learning for skeleton-based action representation seeks to extract discriminative features without labels by crafting supervisory signals that capture action semantics. Early works adopted handcrafted pretext tasks in encoder–decoder schemes, such as corrupted skeleton recovery \cite{zheng2018unsupervised}, full-sequence autoencoding \cite{kundu2019unsupervised, su2020predict}, and pose/view disentanglement before reconstruction \cite{nie2020unsupervised}. Subsequent methods leveraged spatiotemporal cues to construct supervision, such as joint color prediction \cite{yang2021skeleton}, inter-frame motion forecasting \cite{cheng2021hierarchical}, multi-scale displacement estimation \cite{kim2022global}, and motion retargeting for view invariance \cite{yang2024view}.

The field then pivoted to contrastive learning: contrasting augmented skeleton pairs via dual encoders \cite{rao2021augmented}, refined by skeleton-specific transformations and cross-architecture contrast \cite{thoker2021skeleton}. Contextual and latent positive mining \cite{li20213d, zhang2022contrastive, shah2023halp}, multi-modal fusion \cite{mao2022cmd, sun2023unified, weng2024usdrl}, attention to action-critical regions \cite{Hua2023SkeAttnCLR, lin2023actionlet}, and dual-prompted pre-training \cite{zhang2023prompted} further advanced representation quality. 

Building on masked autoencoders, recent works \cite{wu2023skeletonmae, mao2023masked, abdelfattah2024s,gong2025rethinking, sun2025towards} achieve strong semantic encoding via partial observation and reconstruction. Existing approaches rely on Transformer decoders built on self-attention and employ a standard mean-squared-error reconstruction loss. This design, however, struggles to flexibly and efficiently support reconstruction involving larger patches. In contrast, our method introduces an effective mechanism tailored for large patch reconstruction, substantially improving computational efficiency while also significantly enhancing the quality of learned feature representations when combined with our focal reconstruction.

\subsection{Saliency-Guided Learning Methods for Skeleton-Based Action}
In self-supervised skeleton action learning, leveraging priors to guide attention toward salient regions has proven effective.
ActCLR \cite{lin2023actionlet} uses gradients from a contrastive loss to identify potential action regions and explicitly incorporates them into its subsequent contrastive learning process. MAMP \cite{mao2023masked} determines which patches to mask based on motion (\ie, inter-frame displacement). 
Although these methods and our AMR exploit salient regions, they address fundamentally different problems and scenarios: ActCLR’s saliency guidance serves the contrastive learning framework, and MAMP uses it to guide the masking strategy. In contrast, AMR embeds motion guidance deeply within the masked-autoencoder reconstruction process, targeting the reduction of information redundancy in large patch reconstruction and the enhancement of feature discriminability. This task-specific optimization, particularly tuned to the challenges introduced by large patches, is the defining innovation that sets AMR apart from existing methods.

\section{Method}
\label{sec:method}

\begin{figure*}[tb!]
\centering\includegraphics[width=2.0\columnwidth]{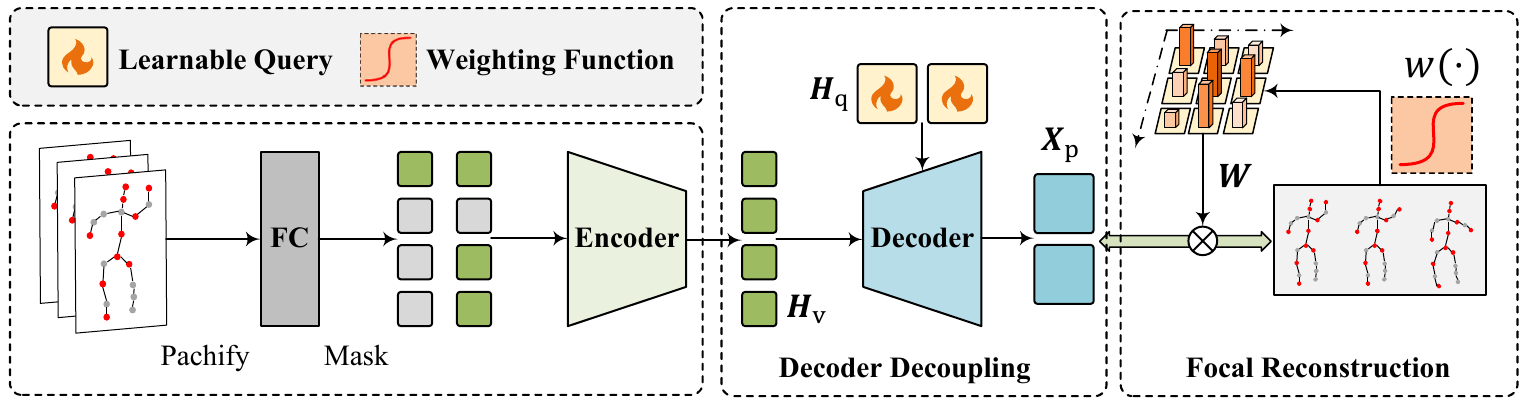} \vspace{-2mm}
\caption{
The framework of our proposed AMR. 
}
\label{fig:framework}
\end{figure*}

We first introduce the pipeline of the masked skeleton reconstruction in Section \ref{sec:method_3_1}. Then, in Section \ref{sec:method_3_2}, we present the decoupled encoder-decoder design to enable flexible and efficient prediction of larger patches. Finally, in Section \ref{sec:method_3_3}, we describe the proposed focal reconstruction mechanism. This mechanism leverages an adaptive guidance strategy to effectively direct the model's focus towards discriminative motion patterns.

\subsection{Preliminaries}
\label{sec:method_3_1}
The standard pipeline of masked skeleton reconstruction proceeds as follows: A predefined masking strategy randomly masks joints, aiming to train the network to reconstruct the masked ones for representation learning.

\textbf{Input.} The input skeleton action sequence is represented as $\mathbf{X} \in \mathbb{R}^{T \times V \times C}$, where $T$ denotes the number of frames, $V$ represents the number of body joints, and $C$ is the number of channels.

\textbf{Patchify and Mask.} To adapt the sequence for the Transformer encoder, it is patchified into spatiotemporal patches. Specifically, continuous $l$ frames of data for each joint are grouped to form a patch, yielding an initial patch representation $\mathbf{X}' \in \mathbb{R}^{T_\text{e} \times V \times (l \times C)}$, where $T_\text{e}=T/l$. A fully connected layer subsequently embeds $\mathbf{X}'$ into a higher-dimensional feature space, which is then flattened to $\mathbf{E} \in \mathbb{R}^{N \times C_\text{e}}$, where $N=T_e \times V$. Then, the majority of patches are randomly masked, retaining only a small subset as visible patches $\mathbf{E}_\text{v} \in \mathbb{R}^{N_\text{v} \times C_\text{e}}$.

\textbf{Encoder.} The Transformer encoder processes only the visible patches $\mathbf{E}_\text{v}$, producing feature embeddings $\mathbf{H}_\text{v} \in \mathbb{R}^{N_\text{v} \times C_\text{e}}$. Since the number of visible patches is small (typically 10\% of the total patches), the computational overhead of the encoder remains relatively low.

\textbf{Decoder.} The objective of the decoder is to reconstruct the original data for the masked patches. Conventional self-attention-based Transformer decoders operate as follows: A set of learnable mask tokens $\mathbf{M}$ is introduced to indicate the presence of a missing patch to be predicted. The visible patch features $\mathbf{H}_\text{v}$ and the mask tokens $\mathbf{M}$ are concatenated to form the decoder input $\mathbf{H}_\text{e}=[\mathbf{H}_\text{v} \| \mathbf{M}] \in \mathbb{R}^{N \times C_\text{e}}$. Notably, the number of obtained patches of the skeleton sequence can be very large. Such a long input sequence leads to a high computational cost for the self-attention module in the decoder. Moreover, unlike the lightweight decoders commonly used in masked image reconstruction methods, the decoder in masked skeleton reconstruction is typically heavier. Consequently, the majority of the training compute budget is consumed during decoding.

\textbf{Reconstruction.} Following the decoder's output of the predicted values $\mathbf{X}_\text{p}\in \mathbb{R}^{N \times (l\times C)}$, the model is optimized by minimizing the mean squared error (MSE) loss between the targets and their reconstructed counterparts:

\begin{equation}
      \mathcal{L} = \frac{1}{N}  \|\mathbf{X}_\text{p} - \mathbf{X}_\text{t}\|_F^2,
\end{equation}
where $\mathbf{X}_\text{t}\in \mathbb{R}^{N \times (l\times C)}$ denotes the reconstruction target, typically obtained by patchifying the original input or its transformed variant.
Under this standard reconstruction scheme, all spatiotemporal regions are treated equally without considering differences in their semantic importance.

\subsection{Decoder Decoupling}
\label{sec:method_3_2}
The low training efficiency in masked skeleton reconstruction primarily stems from the significant computational overhead of the decoder's self-attention mechanism when processing long sequences. Consequently, an intuitive optimization strategy involves reducing the decoder sequence length, which corresponds to increasing the size of target patches during reconstruction (\ie, predicting fewer but larger patches). However, within the standard masked skeleton reconstruction architecture, the self-attention-based decoder is tightly coupled with the encoder. This tight coupling prevents the model from flexibly adjusting the granularity and quantity of prediction targets (\ie, altering reconstructed patch size) while keeping the encoder output $\mathbf{H}_\text{v}$ unchanged. A potential solution is to apply downsampling to the decoder input $\mathbf{H}_\text{e}$ to match the reduced number of target patches. Yet, this forced compression of information inevitably incurs semantic information loss, significantly degrading model performance.

\textbf{Cross-Attention-Based Decoder.} To address this limitation, we re-examine the decoding process. The decoder computation during masked patch prediction comprises two primary components: 1) self-attention computation among mask tokens, and 2) cross-attention computation between mask tokens and the visible patch features $\mathbf{H}_\text{v}$. Crucially, randomly initialized learnable mask tokens inherently lack semantic information. 
We empirically observe that self-attention computation among these tokens is not only computationally expensive but largely unnecessary (Table \ref{tab:decoder_design}). Based on this insight, we simplify the decoder into a lighter, more flexible structure consisting of: (1) a cross-attention layer to fuse information from visible patches, followed by (2) a standard feed-forward network (FFN) layer. This design eliminates the redundant self-attention computation among mask tokens entirely.

As shown in Figure \ref{fig:framework}, the decoder takes a set of learnable query vectors $\mathbf{H}_{\text{q}} \in \mathbb{R}^{N_{\text{t}} \times C_{\text{e}}}$ as input, while utilizing the visible patch features $\mathbf{H}_{\text{v}}$ as both the keys and values. Moreover, learnable spatiotemporal positional encodings are integrated into the queries. The decoding computation comprises the two steps:
\begin{equation}
\begin{split}
\mathbf{H}_{\text{q}} 
&\leftarrow \operatorname{MCA}\left( \mathbf{H}_{\text{q}}, \mathbf{H}_{\text{v}} \right) 
+ \mathbf{H}_{\text{q}}  \\  
\mathbf{H}_{\text{q}} 
&\leftarrow \operatorname{FFN}\left( \mathbf{H}_{\text{q}} \right) 
+ \mathbf{H}_{\text{q}} 
\end{split}
\end{equation}
where $\operatorname{MCA}(\cdot)$ denotes the multi-head cross-attention operation, $\operatorname{FFN}(\cdot)$ is a feedforward network composed of MLPs. 
Our decoder similarly supports multi-layer stacking, analogous to self-attention-based Transformer decoders. Specifically, the output $\mathbf{H}_{\text{q}}$ from the current layer and the visible patch features $\mathbf{H}_{\text{v}}$ are fed into the subsequent decoder layer for further processing. 

\textbf{Larger Patch Reconstruction.} A key advantage of this architecture is its flexibility: To adjust prediction target granularity (\ie, patch size), we only need to modify the number of queries $\mathbf{H}_{\text{q}}$ to match the count of target patches. This requires no modification to the encoder's output or structure, effectively decoupling the encoder and decoder. 
Formally, the model's current reconstruction target is defined as predicting larger patches scaled to $r$ times the size of the original spatiotemporal patches (\ie, covers $r$ times more frames than a standard patch). Consequently, the target patches are represented as $\mathbf{X}_\text{t} \in \mathbb{R}^{(N/r)\times (r \times l \times C)}$, where $N/r$ is the number of target patches. Correspondingly, the number of query vectors $N_{\text{t}}$ in the decoder is directly set to $N/r$ to generate dimensionally-matched predictions $\mathbf{X}_\text{p} \in \mathbb{R}^{(N/r)\times (r \times l \times C)}$. The increased size of target patches directly reduces the decoder sequence length, thereby significantly decreasing the model's overall computational overhead.

\begin{table*}[t]
\centering
{
\caption{Comparisons with the state-of-the-art masked skeleton reconstruction methods on NTU-60 and PKU-II under linear evaluation protocol. All pre-training runs were completed under identical hardware conditions using a single L20 GPU.
}
\label{table:ntu60_linear_eval}
}

\begin{tabular}{@{}l* {8}c l@{}}
\toprule
 \multirow{2}{*}{\textbf{Method}}  &
\multirow{2}{*}{\textbf{Patches}}  &
\multirow{2}{*}{\textbf{FLOPs}}   &
\multirow{2}{*}{\textbf{Hours}}   &
\multirow{2}{*}{\textbf{Speedup}}   &
\multicolumn{2}{c}{\textbf{NTU-60}} &
\multicolumn{1}{c}{\textbf{PKU-II}} 
\\
 \cmidrule(r){6-7} \cmidrule(r){8-8}
&&&&& x-sub & x-view & x-sub \\
\midrule 
SkeletonMAE \cite{wu2023skeletonmae} & 750 & 13.7G  & 29.9h & 1$\times$ & 74.8 & 77.7 & 36.1 \\
MAMP \cite{mao2023masked} & 750 & 13.7G  & 29.9h & 1$\times$ & 84.9 & 89.1 & 53.8 \\
S-JEPA \cite{abdelfattah2024s} & 750 & 32.8G  & 139.8h & 0.2$\times$ & 85.3 & 89.8 & 53.5\\
GFP \cite{sun2025towards} & 251 & 4.1G & 4.9h & 6.1$\times$ & 85.9 & 92.0 & 56.2 \\
NAT \cite{gong2025rethinking} & 750 & 13.7G  & 29.9h & 1$\times$ & 85.1 & - & - \\
NAT w/ Con \cite{gong2025rethinking} & 750 & 32.8G  & 46.6h & 0.6$\times$ & 86.9 & 91.0 & 55.3 \\
\rowcolor{deemph}
AMR (Ours) & \textbf{125} & \textbf{3.8G}  & \textbf{3.7h} & \textbf{8.1$\times$} & \textbf{87.4} & \textbf{92.3} & \textbf{60.3} \\

\bottomrule
\end{tabular}

\vspace{-0.4cm}
\end{table*}

\subsection{Focal Reconstruction}
\label{sec:method_3_3}
After enabling flexible reconstruction of larger patches, we expect its performance to approach that of small patch reconstruction, thereby achieving a favorable trade-off between computational efficiency and model performance. The primary cause of performance degradation lies in the increased difficulty of reconstructing large patches: compared to small patches, large patches contain more complex and redundant spatiotemporal information, making it considerably harder for the model to precisely capture discriminative motion semantics.

\textbf{Weight Assignment.} To mitigate this issue, we introduce a simple yet effective adaptive guided mechanism. The core idea of this mechanism is to quantify the motion energy within each region, derive its corresponding reconstruction weighting coefficient, and thereby guide the model to adaptively reduce attention to redundant information during reconstruction while focusing on those key regions that are critical for understanding action semantics. Typically, discriminative semantic cues in skeleton-based actions are predominantly concentrated in regions of intense motion, whereas static areas tend to carry redundant information. Accordingly, we employ local motion energy as the metric to dynamically assign importance weights during reconstruction.

In practice, we quantify motion energy per joint using its trajectory within non-overlapping temporal windows. The input skeleton sequence $\mathbf{X} \in \mathbb{R}^{T \times V \times C}$ is partitioned into $T_\text{w} = T/n$ windows along the temporal dimension, each spanning $n$ frames. This yields windowed representations $\mathbf{X}_{\text{w}} \in \mathbb{R}^{T_\text{w} \times n \times V \times C}$. For a joint's window slice $\mathbf{S} \in \mathbb{R}^{n \times C}$, its motion energy $e_{\text{s}}$ is defined as the sum of squared norms of the displacement relative to their mean positions within the window:
\begin{equation}
    e_{\text{s}} = \frac{1}{n} \sum_{t=1}^n {\|\mathbf{s}_t - \bm{\mu} \|_2^2},
\end{equation}
where $\mathbf{s}_t$ is the vector of $t$-th frame in $\mathbf{S}$, and $\bm{\mu} = \frac{1}{n} \sum_{t=1}^n \mathbf{s}_t$. 
Based on the computed motion energy $e_{\text{s}}$, we define a well-behaved weight assignment function $\operatorname{w}(\cdot)$:
\begin{equation}
    \operatorname{w}(e_{\text{s}}) = \frac{1}{1+k \cdot  e_{\text{t}}/ (e_{\text{s}}+\epsilon)} ,
\end{equation}
where $e_{\text{t}}$ is an adaptively computed target energy threshold, set as the average motion energy across all windows in the entire action sequence; $k>0$ controls the function's curvature; and $\epsilon$ denotes a small constant added for numerical stability to prevent division by zero. As defined, $\operatorname{w}(e_{\text{s}}) \xrightarrow{} 0$ when the window's motion energy is significantly lower than the threshold, and $\operatorname{w}(e_{\text{s}}) \xrightarrow{} 1$ when $e_{\text{s}} \gg e_{\text{t}}$. Given the orders-of-magnitude variations in motion energy observed empirically, this weighting function ensures smooth adaptation to dynamic motion intensity ranges.

\textbf{Multi-Scale Fusion.} Methods that quantify joint motion energy using a single temporal window are inevitably limited. Short windows tend to miss the slow-varying details, whereas long windows have difficulty capturing rapid movements. To mitigate these issues, we compute joint motion energy across multiple temporal windows at different scales, providing a more comprehensive representation of each joint’s dynamics. In our experiments, the temporal window scale $n$ is set to 4, 8, and 12.
By fusing information from these different temporal contexts, we derive a joint importance measure that integrates both short- and long-term cues.

\textbf{Adaptive Guidance.} After obtaining the weight $\mathbf{W} \in \mathbb{R}^{T_{\text{w}} \times V \times 1}$ for all windows, we integrate it into the reconstruction loss to guide the model to prioritize modeling critical regions. The weights are broadcast across temporal window frames and feature channels, resulting in $\mathbf{W} \in \mathbb{R}^{T_{\text{w}} \times V \times n \times C}$. Then $\mathbf{W}$ is reshaped to match the exact dimensions of the model's prediction output $\mathbf{X}_\text{p}$. Using this reshaped weight tensor $\mathbf{W}$, we define the focal reconstruction loss as:
\begin{equation}\label{eq:mse}
      \mathcal{L} = \frac{1}{N}  \|\sqrt{\mathbf{W}} \odot (\mathbf{X}_\text{p} - \mathbf{X}_\text{t})\|_F^2,
\end{equation}
where $\odot$ represents element-wise multiplication. This weighting mechanism enables the model to adaptively concentrate on regions containing critical motion semantics. Consequently, it substantially enhances the comprehension efficiency of action semantics within large patches and facilitates learning of more discriminative representations.

\section{Experiments}
\label{sec:experiments}

\subsection{Datasets}
We evaluate our method on three skeleton action benchmarks: NTU RGB+D 60 (NTU-60) \cite{shahroudy2016ntu}, NTU RGB+D 120 (NTU-120) \cite{liu2019ntu}, and PKU-MMD II (PKU-II) \cite{liu2020benchmark}. 




Consistent with recent work~\cite{thoker2021skeleton,mao2022cmd,abdelfattah2024s}, all experiments report \textbf{top-1 accuracy}.

\subsection{Comparison with the State-of-the-art Methods}
To evaluate the performance of the encoder obtained through pre-training, we employ multiple standard evaluation protocols, including linear evaluation, semi-supervised evaluation, and transfer learning. Comprehensive comparisons are conducted against recent state-of-the-art (SOTA) methods across multiple skeleton-based action recognition benchmark datasets.
\begin{table}[t]
\centering
{
\caption{Comparisons with the state-of-the-art methods on NTU-120 under linear
evaluation protocol. 
}
\label{table:ntu120_linear_eval}
}
\scalebox{0.95}{
\begin{tabular}{@{}ll* {2}c l@{}}
\toprule
 \multirow{2}{*}{\textbf{Method}}  &
\multicolumn{2}{c}{\textbf{NTU-120}} 
\\
\cmidrule(r){2-3}
 & x-sub & x-setup  \\
\midrule 
P\&C \cite{su2020predict} & 42.7 & 41.7  \\
AimCLR \cite{guo2022contrastive} & 63.4 & 63.4  \\
HaLP\cite{shah2023halp}  & 71.1 & 72.2  \\
HiCo\cite{dong2023hierarchical}  & 72.8 & 74.1  \\
CMD \cite{mao2022cmd} & 70.3 & 71.5  \\
UmURL\cite{sun2023unified}  & 73.5 & 74.3  \\
HSARL \cite{wang2025heterogeneous} & 73.5 & 78.4 \\
USDRL \cite{weng2024usdrl} & 76.6 & 78.1   \\
\midrule
SkeletonMAE\cite{wu2023skeletonmae} & 72.5 & 73.5  \\
MAMP \cite{mao2023masked} & 78.6 & 79.1  \\
NAT \cite{gong2025rethinking} & 78.8 & - \\
GFP \cite{sun2025towards} & 79.1 & 80.3 \\
S-JEPA \cite{abdelfattah2024s} & 79.6 & 79.9  \\
\rowcolor{deemph}
AMR (Ours) & \textbf{81.1} & \textbf{81.9}  \\
\bottomrule
\end{tabular}
}
\vspace{-0.2cm}
\end{table}

\textbf{Linear Evaluation Results:} Under the linear evaluation protocol, the pre-trained encoder is frozen, with a linear classifier appended. This classifier is trained under supervision for 100 epochs with a batch size of 256 and an initial learning rate of 0.01. The learning rate is gradually reduced to 0 using a cosine decay schedule. As shown in Tables \ref{table:ntu60_linear_eval} and \ref{table:ntu120_linear_eval}, we report the performance of the proposed method on the NTU-60, NTU-120, and PKU-II datasets, comparing it with existing approaches.
In addition to recognition accuracy, we compute the FLOPs required to process one action instance and report the pre-training time under the NTU-60 x-sub protocol. Results in Table \ref{table:ntu60_linear_eval} demonstrate that our method surpasses prior approaches in performance while achieving significant training acceleration. This efficiency gain primarily stems from the substantial reduction in the number of reconstruction targets (125 patches vs. 750 in existing methods).
On the challenging PKU-II dataset, our method achieves notable performance gains over previous methods. This strongly validates the effectiveness of the proposed adaptive focal reconstruction strategy, which successfully guides the model to focus on information more discriminative for understanding action semantics.
Table \ref{table:ntu120_linear_eval} presents results on the larger-scale NTU-120 dataset, extended to include methods beyond masked reconstruction. Our method again outperforms existing approaches, further demonstrating the stronger discriminative power of the learned feature representations.

\begin{table} [tb!]
\caption{Comparisons with the state-of-the-art methods with semi-supervised learning on NTU-60 dataset.
}
\label{tab:semisupervised}
%
\centering 
\scalebox{0.9}{
\begin{tabular}{@{}l*{5}c @{}}
\toprule
\multirow{2}{*}{\textbf{Method}}   &
\multicolumn{2}{c}{\textbf{x-sub}} & \multicolumn{2}{c}{\textbf{x-view}} \\

\cmidrule(r){2-3} \cmidrule(r){4-5} 
& 1\%  & 10\%  & 1\%  & 10\%  \\
\cmidrule{1-5}
ASSL \cite{si2020adversarial} & - & 64.3 & - & 69.8   \\
Hi-TRS \cite{chen2022hierarchically} & - & 70.7 & - & 74.8 \\
HaLP \cite{shah2023halp} & 46.6 & 72.6 & 48.7 & 77.1 \\
CMD \cite{mao2022cmd} & 50.6 & 75.4 & 53.0 & 80.2 \\
CPM \cite{zhang2022contrastive} & 56.7 & 73.0 & 57.5 & 77.1 \\
HYSP \cite{franco2023hyperbolic} & - & 76.2 & - & 80.4  \\
HiCo \cite{dong2023hierarchical} & 54.4 & 73.0 & 54.8 & 78.3 \\
UmURL \cite{sun2023unified} & 58.1 & - & 58.3 & - & \\
HSARL \cite{wang2025heterogeneous} & 55.0 & - & 55.0 & - \\
USDRL \cite{weng2024usdrl} & 57.3 & 80.2 & 60.7 & 84.0 \\
\cmidrule{1-5}
SkeletonMAE \cite{wu2023skeletonmae} & 54.4 & 80.6 & 54.6 & 83.5 \\
MAMP \cite{mao2023masked} & 66.0 & 88.0 & 68.7 & 91.5 \\
S-JEPA \cite{abdelfattah2024s} & 67.5 & 88.4 & 69.1 & 91.4 \\
NAT \cite{gong2025rethinking} & 66.2 & 88.3 & 68.8 & 91.8 \\
GFP \cite{sun2025towards} &71.8 & 88.7 & 72.9 & 92.1 \\
\rowcolor{deemph}
AMR (Ours) & \textbf{72.2} & \textbf{89.0} & \textbf{74.4} & \textbf{92.7} \\
\bottomrule
\end{tabular}
 }
\vspace{-0.2cm}
\end{table}
\textbf{Semi-supervised Evaluation Results:} Following the setup of prior studies \cite{thoker2021skeleton,dong2023hierarchical}, under the semi-supervised evaluation protocol, both the pre-trained encoder and the classifier undergo end-to-end fine-tuning using a small fraction of labeled training data. Consistent with \cite{mao2023masked,gong2025rethinking}, we report recognition accuracy on the NTU-60 dataset with only 1\% and 10\% labeled training data. As shown in Table \ref{tab:semisupervised}, our proposed AMR method significantly outperforms recent approaches like USDRL \cite{weng2024usdrl}, and HSARL \cite{wang2025heterogeneous}. Particularly in the extremely low-resource setting with merely 1\% labeled data, AMR demonstrates its most pronounced performance advantage. This result demonstrates that the representations learned by AMR exhibit superior generalization capability, effectively transferring and capturing essential semantic characteristics of actions even when fine-tuned with minimal labeled samples.

\begin{table} [tb!]
\centering 
\caption{Comparisons with the state-of-the-art methods under the transfer learning setting.}
\vspace{-2mm}
\label{tab:transfer}
\scalebox{0.95}{
\begin{tabular}{@{}l*{3}c @{}}
\toprule
\multirow{2}{*}{\textbf{Method}}   & 
\multicolumn{2}{c}{\textbf{To PKU-II}} &\\
\cmidrule{2-3} 
& \multicolumn{1}{c}{\textbf{NTU-60}} &
\multicolumn{1}{c}{\textbf{NTU-120}} &\\
\hline
M$^2$L \cite{lin2020ms2l}  & 45.8 & - \\
CrosSCLR \cite{li20213d}  & 54.0 & 52.8 \\
HiCo \cite{dong2023hierarchical}  & 56.3 & 55.4 \\
CMD \cite{mao2022cmd}  & 56.0 & 57.0 \\
UmURL \cite{sun2023unified}  & 59.7 & 58.5 \\ 
HSARL \cite{wang2025heterogeneous} & 64.3 & 63.1 \\
\midrule
SkeletonMAE \cite{wu2023skeletonmae} & 58.4 & 61.0\\
MAMP \cite{mao2023masked} & 70.6 & 73.2 \\
NAT \cite{gong2025rethinking} & 70.7 & \textbf{73.2} \\
\rowcolor{deemph}
AMR (Ours) & \textbf{70.9} & 73.0 \\
\bottomrule
\end{tabular}
}
\vspace{-2mm}
\end{table}
\textbf{Transfer Learning Evaluation Results:} 
Following \cite{mao2022cmd,sun2023unified}, under the transfer learning evaluation protocol, the model is first pre-trained on a source dataset and subsequently fine-tuned on a target dataset. This protocol is designed to assess the generalization ability of representations learned during pre-training. In our experiments, PKU-II serves as the target dataset, with NTU-60 and NTU-120 as source datasets. As evidenced by the results in Table \ref{tab:transfer}, AMR demonstrates strong competitiveness compared to existing methods: when transferred to PKU-II after pre-training on different source datasets, AMR consistently matches or surpasses SOTA performance. This strongly confirms the superior transferability of features learned by AMR.

\subsection{Ablation Study}
All ablation studies in this section are conducted under either the NTU-60 x-sub or NTU-120 x-sub protocol. We first validate the effectiveness of two core components: the decoupled decoder and focal reconstruction mechanism. Subsequently, we perform an in-depth analysis of critical design choices and hyperparameters within each module.

\begin{table} [tb!]
\renewcommand{\arraystretch}{1.2}
\centering 
\caption{Ablation study on focal reconstruction and decoder decoupling.}
\vspace{-2mm}
\label{tab:ablation_fr_dd}
\scalebox{0.95}{
\begin{tabular}{l*{3}c @{}}
\toprule
\multicolumn{1}{l}{\textbf{Method}} & \multicolumn{1}{c}{\textbf{NTU-60}} &
\multicolumn{1}{c}{\textbf{NTU-120}} \\
\hline
Baseline1 (Decoder Masking) & 84.8 & 76.3 \\
Baseline2 (Downsampling) & 78.9 & 70.2 \\
Baseline + DD & 86.0 & 80.1 \\
Baseline + DD + FR & 87.4 & 81.1 \\
\bottomrule
\end{tabular}
}
\vspace{-3mm}
\end{table}

\begin{figure}[tb!]
\centering
\includegraphics[width=0.85\columnwidth]{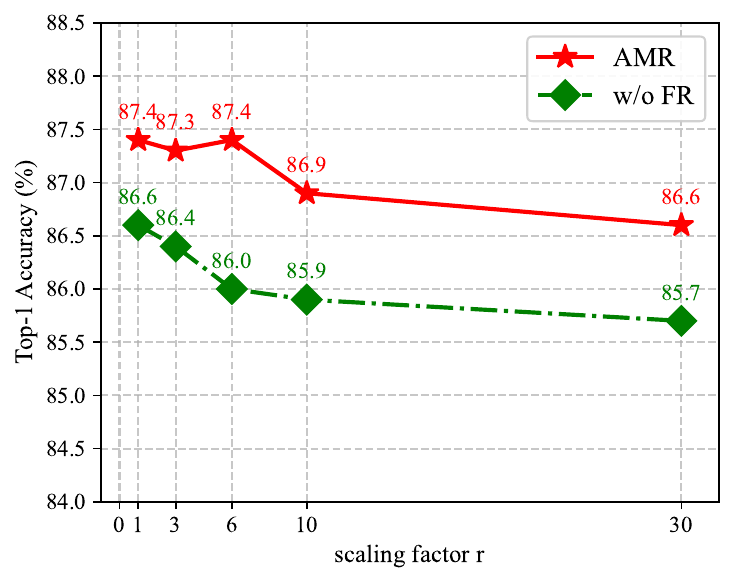}
\vspace{-5mm}
\caption{Impact of patch size on model performance. 
}\label{fig:facor_r}
\vspace{-3mm}
\end{figure}

\begin{figure*}[tb!]
\centering
\includegraphics[width=2.0\columnwidth]{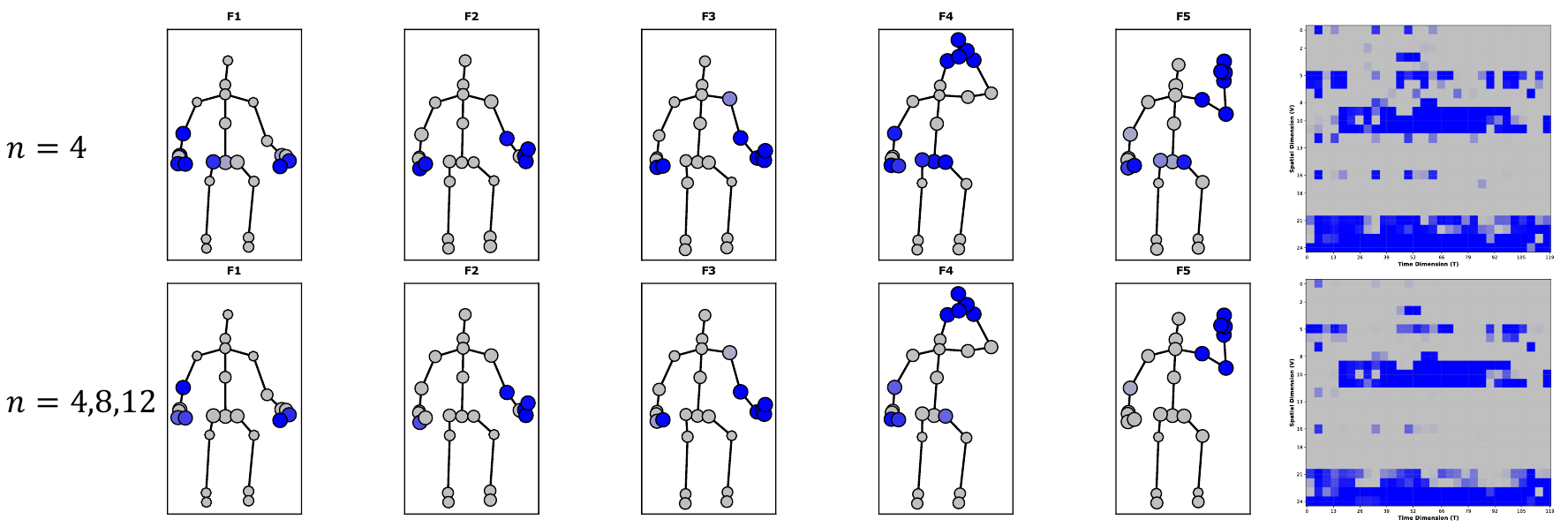}
\vspace{-2mm}
\caption{Visualization of joint importance for the ``take off a hat" action. Joints with high weight are depicted in blue.  
}\label{fig:hyperparam}
\vspace{-4mm}
\end{figure*}

\textbf{Decoupled Decoder and Focal Reconstruction:}
Table \ref{tab:ablation_fr_dd} presents the ablation results for the key components of our method. Under the experimental setting where all models reconstruct 125 patches as targets, we compare two baseline self-attention–based decoder strategies. Baseline 1 selects a reduced set of patches as reconstruction targets via decoder masking \cite{wang2023videomae}; Baseline 2 targets coarser spatio-temporal patches but requires downsampling \cite{sun2025towards} in the decoding stage to match the target count. We replace the original decoding module with our proposed Decoupled Decoder (\textbf{DD}) module and evaluate the effects. The results show that, when reconstructing larger patches, the original decoding strategies suffer a substantial performance drop. This degradation is mainly caused by the tight coupling between the standard decoder and encoder, together with the information loss introduced by target masking and downsampling. By contrast, our cross-attention–based decoder, with its flexible query mechanism, effectively mitigates these issues and better preserves feature quality. Adding the Focal Reconstruction (\textbf{FR}) module further improves performance, corroborating its effectiveness in steering the model to concentrate on critical regions.

\textbf{Larger Patch Size:} We further investigate performance trends under varying patch scaling factors $r$, where reconstructed patches are $r$ times larger than standard patches. Figure \ref{fig:facor_r} illustrates performance changes as $r$ increases from 1 (standard size) to 30. Notably, larger $r$ values (larger patches) reduce the number of prediction targets and accelerate training. As shown in Figure \ref{fig:facor_r}, 
the AMR variant w/o FR (green curve), while underperforming full AMR due to lacking focal guidance, significantly surpasses the baseline decode methods in Table \ref{tab:ablation_fr_dd}. This validates the DD module’s inherent effectiveness in adapting to large patch reconstruction. Furthermore, the full AMR (red curve) exhibits exceptional robustness to increasing $r$. Performance remains near peak levels even at $r=6$, achieving a favorable trade-off between training efficiency and model performance. These ablation results provide compelling evidence for the necessity and effectiveness of AMR’s core components. This simple yet efficient architecture establishes a strong baseline for self-supervised skeleton action recognition.

\begin{table} [tb!]
\renewcommand{\arraystretch}{1.1}
\centering 
\caption{Ablation study on decoder. MCA and MSA denote multi-head cross-attention and multi-head self-attention.}
\vspace{-2mm}
\label{tab:decoder_design}
\scalebox{0.95}{
\begin{tabular}{l*{3}c @{}}
\toprule
\multicolumn{1}{l}{\textbf{Method}} & \multicolumn{1}{c}{\textbf{NTU-60}} &
\multicolumn{1}{c}{\textbf{NTU-120}} \\
\hline
MCA & 87.4 & 81.1 \\
MCA+MSA & 87.2 & 81.2 \\

\bottomrule
\end{tabular}
}
\vspace{-2mm}
\end{table}
\textbf{Decoder Design:} As previously established in this work, self-attention computation among mask tokens in the decoder is non-essential for feature learning. To further validate this claim, we explicitly incorporated a self-attention layer after the cross-attention module in the decoder, forcing information exchange between mask tokens. Results in Table \ref{tab:decoder_design} demonstrate that this modification yields no statistically significant performance change. This empirical evidence directly substantiates our earlier analysis: within the skeleton masked autoencoding framework, self-attention among mask tokens is computationally redundant and can be safely removed without compromising representation capability.

\textbf{Visualization:} To validate our proposed weight assignment method, we select a sample of the ``taking off a hat" action for visual analysis. In the illustrations, joints with high weights (\textgreater 0.9) are marked in blue within the skeleton sequence and emphasized in the spatio-temporal weight heatmap on the right. The upper panel shows joint importance computed using a single temporal window, while the lower panel displays the results after multi-scale temporal fusion. The comparison indicates that the multi-scale fusion effectively reduces misjudgments of less discriminative joints, as evidenced by the last two skeleton diagrams. Moreover, the heatmap on the right reveals that single-window weights exhibit several transient high-weight intervals along the temporal axis (x-axis), whereas the multi-window results are more continuous and stable over time. This continuity suggests that multi-scale computation captures action-relevant key joints more consistently and demonstrates improved robustness.

\begin{table} [tb!]
\renewcommand{\arraystretch}{1.1}
\centering 
\caption{Ablation study on hyperparameter settings.}
\label{tab:hyper_param}
\scalebox{1.0}{
\begin{tabular}{l*{3}c @{}}
\toprule
\multicolumn{1}{l}{\textbf{}} & \multicolumn{1}{c}{\textbf{NTU-60}} &
\multicolumn{1}{c}{\textbf{NTU-120}} \\
\hline
$k$=0.001 & 86.9 & 80.7 \\
$k$=0.01 & 87.4 & 81.1 \\
$k$=0.1 & 87.3 & 81.0 \\

\bottomrule
\end{tabular}
}
\vspace{-2mm}
\end{table}
\textbf{Hyperparameters:} We conduct an investigation into the impact of hyperparameters in the focal reconstruction module under the NTU-60 x-sub protocol. Table \ref{tab:hyper_param} presents detailed experimental results. The model performs best at $k = 0.01$.

\section{Conclusion}
\label{sec:conclusion}

This paper proposes AMR, a simple yet efficient self-supervised pre-training framework for skeleton-based action recognition. Its innovations include: (1) A decoupled encoder-decoder architecture that dramatically reduces computational complexity while enabling flexible prediction of variably-sized reconstruction targets; (2) An adaptive focal reconstruction mechanism guided by motion energy priors, which effectively focuses modeling capacity on highly discriminative motion semantics, substantially enhancing feature quality during larger patch reconstruction. Extensive experiments across multiple datasets validate the effectiveness of our proposed method.

\section*{Acknowledgments}
This work was supported by the National Natural Science Foundation of China (No. 72188101, 62020106007, and 62472385), Young Elite Scientists Sponsorship Program by China Association for Science and Technology (No. 2022QNRC001) and Fundamental Research Funds for the Provincial Universities of Zhejiang (No. FR2402ZD).

{
    \small
    \bibliographystyle{ieeenat_fullname}
    \bibliography{main}
}

\clearpage            
\appendix          
\section{Details on Decoder Decoupling}
To decouple the decoder, we introduce a cross-attention module into its design. The following details its computational workflow. Let the input query, key, and value tensors be denoted by $\mathbf{Q} \in \mathbb{R}^{N_{\text{q}} \times C}$, $\mathbf{K} \in \mathbb{R}^{N_{\text{k}} \times C}$, and $\mathbf{V} \in \mathbb{R}^{N_{\text{k}} \times C}$, respectively. The output of this module can be formulated as:
\begin{equation}
    \operatorname{MCA}(\mathbf{Q},\mathbf{K},\mathbf{V})=\operatorname{Concat}({head}_1,\cdots,{head}_h){\mathbf{W}}^O,
\end{equation}
where $\mathbf{W}^O \in \mathbb{R}^{C \times C}$ denotes the projection matrix for the output, ${head}_i$ represents the output of the $i$-th cross-attention head, and $\operatorname{Concat}(\cdot)$ refers to the concatenation operation across all heads. The computation of each cross-attention head is defined as follows:
\begin{equation}
    {head}_i =\operatorname{softmax}(\frac{\mathbf{Q}\mathbf{W}_i^Q(\mathbf{K}\mathbf{W}_i^K)^{\top}}{\sqrt{d_k}}) \mathbf{V}\mathbf{W}_i^V
\end{equation}
where $d_k = C/h$, $\mathbf{W}_i^Q \in \mathbb{R}^{C \times d_k}$, $\mathbf{W}_i^K \in \mathbb{R}^{C \times d_k}$ and $\mathbf{W}_i^V \in \mathbb{R}^{C \times d_k}$ are the parameter matrix of the $i$-th cross-attention head. In our architecture, $\mathbf{H}_{\text{q}}$ serves as the query, while $\mathbf{H}_{\text{v}}$ is utilized as both the key and value in the cross-attention mechanism.

\section{Datasets}
We evaluate on three skeleton action benchmarks: NTU RGB+D 60 (NTU-60), NTU RGB+D 120 (NTU-120), and PKU-MMD II (PKU-II). The following contents present detailed descriptions of the datasets used in our experiments.

\textbf{NTU RGB+D 60}~\cite{shahroudy2016ntu} contains 56,880 sequences of 60 actions from 40 subjects under multi-view camera setups. Standard protocols include:
Cross-subject (x-sub): 20 subjects for training, 20 for testing; Cross-view (x-view): Views 2-3 for training, View 1 for testing

\textbf{NTU RGB+D 120}~\cite{liu2019ntu} extends NTU-60 to 120 actions with 114,480 sequences from 106 subjects across 32 setups. Evaluation uses:
Cross-subject (x-sub): 53 subjects per training/test set; Cross-Setup (x-setup): Even-numbered setups for training, odd for testing

\textbf{PKU-MMD II}~\cite{liu2020benchmark} provides 6,952 sequences of 51 actions with significant viewpoint variations (31 camera viewpoints). Following previous work \cite{sun2023unified,gong2025rethinking}, we adopt the cross-subject split: 5,339 training and 1,613 test samples.

\subsection{Implementation Details}
\textbf{Model Architecture.}  
Following \cite{mao2023masked, abdelfattah2024s}, our model employs a standard Vision Transformer backbone with learnable spatio-temporal positional embeddings. The encoder comprises 8 identical layers, each containing multi-head self-attention (8 heads, 256-dimensional) and a feed-forward network with 1024 hidden units. During pre-training, the cross-attention decoder maintains identical feature dimensions but reduces depth to 5 layers. All decoder outputs are projected to the target dimension via a 3-layer MLP for reconstruction.
\\
\textbf{Data Augmentation.}  
For skeleton sequence processing, as in previous works \cite{mao2023masked, abdelfattah2024s}, we adopt temporal cropping with dynamic scaling by randomly selecting a continuous clip spanning 50\%-100\% of the original sequence length. All clips are resampled to frames via bilinear interpolation to ensure temporal consistency.
\\
\textbf{Pre-training Setup.}  
We employ the motion-aware masking strategy proposed in \cite{mao2023masked} and segment length $l$ to 4 in the pertaining. Optimization uses AdamW ($\beta_1=0.9$, $\beta_2=0.95$, weight decay 0.05) under a 400-epoch schedule. Learning rates ramp up linearly from 0 to 1e-3 during 20 warmup epochs, then decay to 5e-4 via cosine annealing. Hyperparameter $k$ is set to 0.01.


\section{Additional Experiments}
All experiments in this section are conducted under either the NTU-60 x-sub or NTU-120 x-sub protocol.

\begin{table} [tb!]
\renewcommand{\arraystretch}{1.1}
\centering 
\caption{Ablation study on reconstruction target.}
\label{tab:target}
\scalebox{1.0}{
\begin{tabular}{l*{3}c @{}}
\toprule
\multicolumn{1}{l}{\textbf{Target}} & \multicolumn{1}{c}{\textbf{NTU-60}} &
\multicolumn{1}{c}{\textbf{NTU-120}} \\
\hline
Joint & 87.4 & 81.1 \\
Motion & 86.7 & 80.6 \\

\bottomrule
\end{tabular}
}
\end{table}

\subsection{Reconstruction Target}
Regarding reconstruction targets, we follow NAT's configuration \cite{gong2025rethinking} by using temporally normalized raw joint coordinates. For a comprehensive comparison, we additionally adopt MAMP's approach \cite{mao2023masked} where motion (inter-frame displacement) serves as the supervisory signal. Results in Table \ref{tab:target} demonstrate that motion-based targets achieve comparable recognition accuracy to conventional targets. Notably, under same motion target settings, our method surpasses MAMP. This outcome further validates that the AMR framework effectively learns discriminative action representations.

\begin{table} [tb!]
\renewcommand{\arraystretch}{1.1}
\centering 
\caption{Ablation study on loss computation.}
\label{tab:loss_computaion}
\scalebox{1.0}{
\begin{tabular}{l*{3}c @{}}
\toprule
\multicolumn{1}{l}{\textbf{Loss Type}} & \multicolumn{1}{c}{\textbf{NTU-60}} &
\multicolumn{1}{c}{\textbf{NTU-120}} \\
\hline

Only Masked Patches & 87.1 & 80.7 \\
All Patches & 87.4 & 81.1 \\

\bottomrule
\end{tabular}
}
\end{table}

\subsection{Loss on Visible Patches}
Previous methods \cite{mao2023masked,abdelfattah2024s} typically compute the loss only on masked patches, which can yield slightly better performance. In our approach, we also compute the reconstruction loss on visible patches. Under the large patch reconstruction setup, a single large patch may contain both visible and masked patches, and zeroing out the reconstruction loss for the visible portions (as in prior work) would provide an incomplete and fragmented supervisory signal that hinders learning of coherent action representations. Table \ref{tab:loss_computaion} compares these two loss calculation strategies: the results show that, with large patch reconstruction, computing the loss over all targets consistently produces better performance.

\subsection{Training Stability}
Because each sample is assigned a different weight, a few extreme weights may shrink the loss of certain samples excessively, potentially affecting training stability. To ensure that the loss scale of each sample does not fluctuate dramatically, we explore two more stable loss computation strategies. First, we clamp extremely small weights to 0.2 to prevent any sample from being completely ignored. Second, we apply a per-sample normalized weighted mean, which normalizes each sample’s loss to make them comparable across the batch. As shown in Table \ref{tab:stability}, all three strategies achieve similar performance, indicating that the original method is already sufficiently stable during training.

\begin{table} [tb!]
\renewcommand{\arraystretch}{1.1}
\centering 
\caption{Ablation study on weight assignment strategy that affects training stability.}
\label{tab:stability}
\scalebox{1.0}{
\begin{tabular}{l*{3}c @{}}
\toprule
\multicolumn{1}{l}{\textbf{Strategy}} & \multicolumn{1}{c}{\textbf{NTU-60}} &
\multicolumn{1}{c}{\textbf{NTU-120}} \\
\hline
Clamp(min=0.2) & 87.3 & 81.2 \\
Normalization  & 87.4 & 80.9 \\
Original & 87.4 & 81.1 \\

\bottomrule
\end{tabular}
}
\end{table}

\begin{table} [tb!]
\renewcommand{\arraystretch}{1.1}
\centering 
\caption{Ablation study on reconstruction target.}
\label{tab:decoder_depth}
\scalebox{1.0}{
\begin{tabular}{l*{3}c @{}}
\toprule
\multicolumn{1}{l}{\textbf{Depth}} & \multicolumn{1}{c}{\textbf{NTU-60}} &
\multicolumn{1}{c}{\textbf{NTU-120}} \\
\hline
4 & 86.8 & 80.7 \\
5 & 87.4 & 81.1 \\
6 & 87.2 & 80.8 \\

\bottomrule
\end{tabular}
}
\end{table}

\subsection{Decoder Depth}
To maintain comparability with prior methods, our decoder's default depth is similarly set to 5 layers. We further investigated depth impact through ablation studies with varied layer counts. As shown in Table \ref{tab:decoder_depth}: the 5-layer configuration consistently produces optimal performance.

\end{document}